\documentclass[11pt]{article}
\usepackage[margin=0.92in]{geometry}
\usepackage{amsmath,amssymb,amsthm}
\usepackage{booktabs,tabularx,array}
\usepackage{xcolor,tikz,pgfplots}
\pgfplotsset{compat=1.18}
\usetikzlibrary{arrows.meta,positioning}
\usepackage{enumitem}
\usepackage[numbers,sort&compress]{natbib}
\usepackage[hidelinks]{hyperref}
\usepackage{url}
\setlist{nosep,leftmargin=*}
\newcolumntype{Y}{>{\raggedright\arraybackslash}X}
\newtheorem{proposition}{Proposition}
\newtheorem{corollary}{Corollary}
\newcommand{\Bee}{\textbf{BeeNet}}
\newcommand{\OIM}{\textbf{Open Inference Market}}
\title{A Compute Center Anyone Can Join:\\A Peer-to-Peer System for AI Inference}
\author{Yimo Lin$^{1}$\thanks{Correspondence: \texttt{murdmonoto@gmail.com}}\quad Zhen Zhang$^{2}$\\[3pt]
  \small $^{1}$Yunyong Century (Beijing) Artificial Intelligence Technology Co., Ltd.\\
  \small $^{2}$Tsinghua University}
\date{October 2026}

\begin{document}
\maketitle

\begin{abstract}
Today, most people obtain AI inference by buying it from a small number of organizations that run large data centers, while many GPUs owned by individuals and small institutions sit idle for much of the day. Bitcoin showed that strangers can contribute resources under shared rules and receive checkable rewards, but it relies on an asymmetry: proof-of-work is expensive to produce and cheap to verify. Useful inference has no such asymmetry. We propose the \OIM{} (OIM), which replaces the missing asymmetry with an economic mechanism. A worker delivers a result without knowing whether it will be audited; its unsettled earnings serve as collateral during a challenge window; and confirmed cheating forfeits all of it. We call this principle earnings as collateral. In a simplified model we show three things. The number of audits needed for deterrence is a constant per worker per day, independent of the worker's throughput, so the audited share falls as the network gets busier. A new worker can join with a short period of intensive auditing and a very small bond. And whether the mechanism works at all depends only on the ratio of the audit judge's false-flag rate to its detection rate. Under example parameters, about 0.3 replays per worker per day are enough to make cheating unprofitable for a rational worker. On this basis we lay out a path that hands four powers back to participants one at a time: owning compute, writing history, judging work, and releasing value. A compute center need not be a building; it can be a protocol that anyone can join.
\end{abstract}

\section{Introduction: a compute center can be a protocol}
\label{sec:intro}

AI is becoming a basic capability: writing, programming, search, and analysis increasingly begin with a model inference. Yet today there is almost only one answer to where inference comes from: buy it from a few organizations that operate large data centers. Prices, quotas, available models, and terms of service are set by the provider. For individual developers, small teams, and schools and firms in regions with limited resources, access to compute decides how far they can take part in this technological shift.

Meanwhile, compute is already spread across many hands. A designer's workstation, a gamer's PC, a lab's training cards, and a small company's server all sit idle for long stretches of the day. Benkler calls such privately owned goods with natural excess capacity \emph{shareable goods}, and argues that their excess capacity can be organized into production through social sharing rather than only through conventional markets~\citep{benkler2004}. Volunteer computing has already shown the scale: SETI@home, Folding@home, and BOINC joined millions of ordinary computers into scientific computing platforms~\citep{setiathome2002,folding2000,boinc2004}. Today the conditions have matured together for the first time. Open-weight models can complete a wide range of practical tasks; inference and offloading systems for consumer GPUs let a single home card run mid-sized models~\citep{powerinfer2023,flexgen2023}; and efficient serving frameworks and collaborative inference systems show that dispersed devices can offer stable service~\citep{vllm2023,borzunov2023}.

What is missing is trust. Volunteer computing pays nothing, so there is no motive to cheat. Once there is payment, a worker has reason to cut corners, for example by passing off a smaller model as the one requested. This is not hypothetical: deceptive model claims have been documented in paid API markets~\citep{zhang2026shadow}, and auditing model substitution has become a research problem in its own right~\citep{cai2025substitution}. Bitcoin lets strangers contribute compute safely because its work is hard to do and easy to check~\citep{nakamoto2008}. Inference is the opposite: checking an answer costs about as much as computing it again.

Our answer is that we do not need every inference to be cheaply provable; we only need cheating to be unprofitable in expectation. We propose the \OIM{} (OIM), which turns an inference into a transaction with a quote, a commitment, a receipt, a possible audit, an appeal, and delayed settlement, and which rests its security on \textbf{earnings as collateral}. Our contributions are:
\begin{enumerate}[label=(\arabic*)]
\item a job protocol from quote to settlement, whose roles are logically separable, can be held by one operator at first, and can be handed over later (Sections~\ref{sec:protocol} and~\ref{sec:audit});
\item a quantitative analysis of earnings as collateral: a per-job audit rule with a deterrence proof, an audit load independent of throughput, a logarithmic cold-start cost, and a feasibility criterion that depends only on the quality of the audit judge (Section~\ref{sec:econ});
\item an account of how the network starts and grows: its first market, two network effects that strengthen with scale, and designs for availability and privacy (Section~\ref{sec:growth});
\item a path that hands four powers (owning compute, writing history, judging work, releasing value) back to participants one at a time, each stage with publicly checkable milestones (Section~\ref{sec:path}).
\end{enumerate}
Assumptions and open problems are collected in Section~\ref{sec:open}; a live early instance, \Bee{}, is described in Appendix~\ref{app:beenet}.

\section{From hashes to inference}
\label{sec:hash}

Bitcoin's design can be split into three elements: \emph{open contribution}, so that anyone can join without permission; \emph{a checkable record}, so that anyone can inspect the ledger; and \emph{aligned incentives}, so that honest contribution pays better than cheating~\citep{nakamoto2008}. What binds them is the asymmetry of proof-of-work: finding a qualifying hash takes a great deal of computation, while checking it takes a single hash~\citep{dworknaor1992,back2002hashcash}. Because verification is cheap, every participant can check the entire history independently.

Proofs of useful work try to make this effort useful at the same time~\citep{ball2017pouw,ofelimos2022}, but useful work rarely has the property of being hard to do and easy to check. Large language model inference is a clear case: re-verifying an answer almost requires running the model again, and two honest runs on different hardware or batches may still differ~\citep{atil2024nondet,he2025nondet}.

The key step in OIM is to lower the requirement. Bitcoin uses work to maintain a globally agreed ledger, so every unit of work must be verifiable. An inference market only needs \emph{each purchased job to be honest in expectation}. That is an economic problem, and economics has a mature answer: a rational agent does not cheat when the probability of being caught times the penalty exceeds the gain~\citep{becker1968}. Volunteer computing has long used random spot checks against saboteurs~\citep{sarmenta2002,golle2001}. To make this logic work among strangers, three things are needed:
\begin{itemize}
\item \textbf{Unpredictability}: when a worker delivers a result, it cannot know whether this job will be audited.
\item \textbf{Something to lose}: the worker holds value that it will certainly lose if caught.
\item \textbf{Adjudication}: there is a judging procedure, published in advance and open to appeal.
\end{itemize}
In OIM, the something to lose comes from the worker itself. Each job's payment first enters a confirmation window, during which it serves as collateral. The busier a worker is, the more collateral it holds, and no one has to put up capital in advance. Bitcoin entrusts security to a majority of hash power; OIM entrusts it to the cost of being caught.

Nor is this an entirely new social invention. Ostrom studied commons that have endured for long periods and distilled a set of shared design principles, among them clearly defined boundaries, monitors accountable to the users, graduated sanctions, low-cost conflict resolution, and nested layers of organization~\citep{ostrom1990}. OIM writes these principles into a protocol: admission tests draw the boundary, audits and public records provide monitoring, penalties escalate from quarantine to forfeiture, appeals resolve disputes, and multiple operators sharing one rule set form a nested structure. Blockchains let strangers jointly maintain a public ledger; OIM aims to let them jointly provide public compute.

\section{The life of a job}
\label{sec:protocol}

OIM has five roles: a \emph{requester} submits a job; a \emph{gateway} matches and dispatches it; a \emph{worker} is a machine owner contributing a GPU; an \emph{auditor} performs checks; and a \emph{settler} releases payment after the challenge window. Home workers need no inbound ports: each opens an outbound connection to a relay, which forwards authenticated traffic. Every participant signs with a public key. Because new keys can be generated at will, an identity has no value in itself; value comes from the collateral and reputation described below.

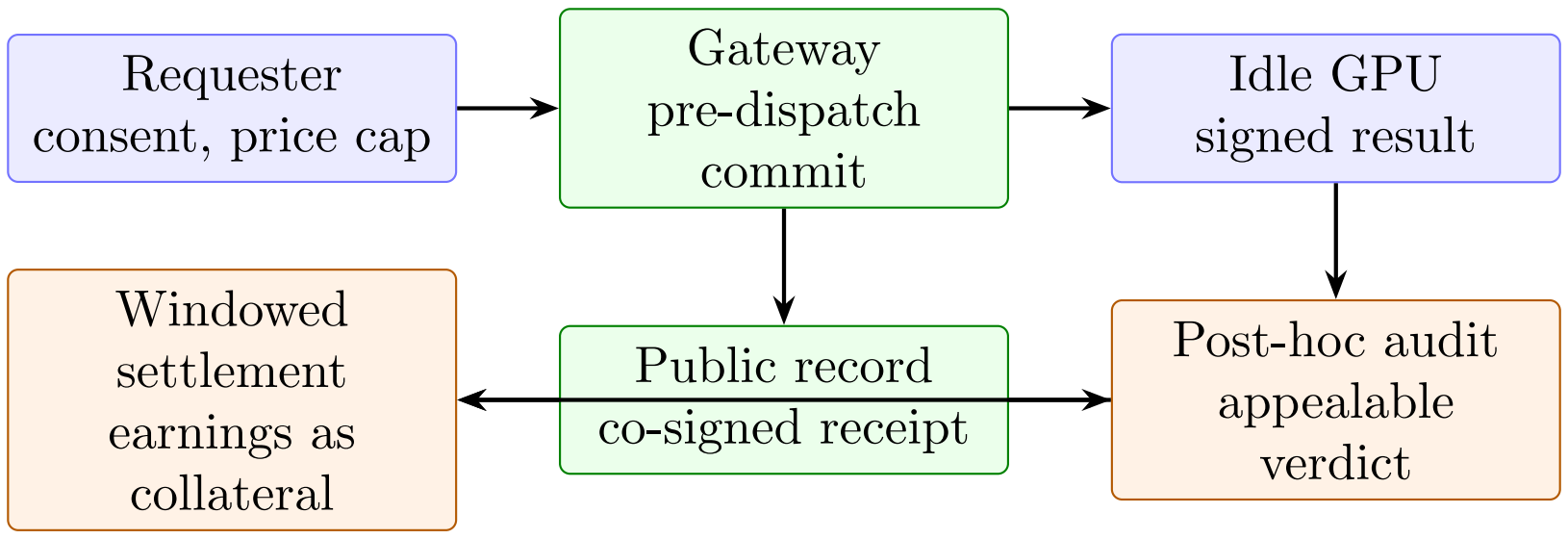
\begin{figure}[t]
\centering
\begin{tikzpicture}[node distance=5mm and 7mm, every node/.style={font=\small},
 box/.style={draw,rounded corners=2pt,align=center,minimum height=9mm,inner sep=4pt},
 arr/.style={-{Stealth[length=2mm]},thick}]
 \node[box,fill=blue!8,draw=blue!55,text width=28mm] (request) {Requester\\consent, price cap};
 \node[box,fill=green!8,draw=green!50!black,right=of request,text width=28mm] (gateway) {Gateway\\pre-dispatch commit};
 \node[box,fill=blue!8,draw=blue!55,right=of gateway,text width=28mm] (worker) {Idle GPU\\signed result};
 \node[box,fill=green!8,draw=green!50!black,below=8mm of gateway,text width=28mm] (log) {Public record\\co-signed receipt};
 \node[box,fill=orange!10,draw=orange!70!black,right=of log,text width=28mm] (audit) {Post-hoc audit\\appealable verdict};
 \node[box,fill=orange!10,draw=orange!70!black,left=of log,text width=28mm] (settle) {Windowed settlement\\earnings as collateral};
 \draw[arr] (request)--(gateway);
 \draw[arr] (gateway)--(worker);
 \draw[arr] (worker)--(audit);
 \draw[arr] (gateway)--(log);
 \draw[arr] (log)--(audit);
 \draw[arr] (audit)--(settle);
 \draw[arr] (log)--(settle);
\end{tikzpicture}
\caption{One OIM transaction. Roles are logically separate; one operator may hold several at first and hand them over as in Section~\ref{sec:path}. Receipts make every job accountable; audits selected only after results are submitted prevent cheaters from timing their behaviour; and the confirmation window turns unsettled earnings into collateral.}
\label{fig:transaction}
\end{figure}

\begin{enumerate}[label=\textbf{\arabic*.}]
\item \textbf{Quote and admission.} A worker signs a quote stating its hardware class, runnable models, capacity, open hours, and price. On admission the gateway checks, with known-answer tests and disconnect-recovery tests, that the worker can honour its quote. Admission tests establish capability; earnings as collateral governs behaviour afterwards.
\item \textbf{Request and consent.} The requester specifies the model, decoding parameters, deadline, price cap, and data tier. Initially, external workers receive only jobs whose requester has explicitly allowed outside machines to read them. The gateway matches on whether a model is already loaded, deadline, price, and risk budget.
\item \textbf{Pre-dispatch commitment.} The gateway signs a unique job envelope containing the chosen quote, model weight digest, runtime and tokenizer versions, parameters, usage cap, price, deadline, a salted input digest, and the round of public randomness that will select audits for this job. Requester and worker each keep a copy, so the gateway cannot quietly rewrite the agreement later.
\item \textbf{Execution and receipt.} The worker returns the output with a signed completion proof; worker and gateway then co-sign a receipt binding job, output digest, usage, and amount due. Every accepted job ends either with a receipt or with an abort record carrying a refund rule. If the gateway refuses to co-sign, the worker can raise a dispute with the envelope and completion proof.
\item \textbf{Audit and settlement.} After the receipt is logged, a random value published only then decides whether the job is audited (Section~\ref{sec:audit}). Payment enters the confirmation window and is released if no dispute arises; confirmed cheating triggers the forfeiture agreed at admission.
\end{enumerate}

The public record holds commitments, receipts, and accounts, not prompts or outputs. When an audit needs the original input, the service retains it under restricted access during the challenge window, deletes it afterwards, and discloses this arrangement to users.

\section{Post-hoc audits and audit contracts}
\label{sec:audit}

An open network must answer three questions, and OIM answers each with a different tool. \emph{Do both sides acknowledge the transaction?} Signed job envelopes and receipts answer this. \emph{Can the operator hide or rewrite records?} An append-only public log with inclusion and consistency proofs answers this, as in Certificate Transparency~\citep{laurie2013ct}. \emph{Did the worker compute as agreed?} Audits and a judge answer this. Each tool does its own job, and none substitutes for another.

Audits must be unpredictable at delivery time. For receipt $R_j$, a public random value $\rho_{t(j)}$ whose round is fixed in the job envelope but which is published only after the receipt-logging deadline, and a published audit rate $p_j$, the rule is
\begin{equation}\label{eq:select}
 \frac{\operatorname{int}\bigl(H(\texttt{OIM-audit-v1}\parallel H(R_j)\parallel\rho_{t(j)})\bigr)}{2^{256}}<p_j .
\end{equation}
Anyone can recompute this selection for every logged receipt, so the operator cannot secretly audit less, audit more, or audit only the workers it chooses. A missing receipt does not become a paid job that happened not to be selected; it enters the abort or dispute process.

A selected job goes to an auditor, who judges it under the model's \emph{audit contract}: the contract states the model weight digest, runtime, decoding settings, eligible hardware classes, comparison method, evidence retention period, and appeal steps. For a fixed hardware class and greedy decoding, replay followed by token-by-token comparison is a simple judge; for sampled decoding and heterogeneous hardware, activation fingerprints, distributional comparison, and cryptographic proofs are advancing quickly~\citep{toploc2025,difr2025,zkllm2024}. A judge's quality is summarized by two numbers: the \emph{detection rate} $d$, the probability that an audited cheat is judged a cheat, and the \emph{false-flag rate} $f$, the probability that an audited honest job is judged a cheat. A judge may abstain on differences it cannot resolve instead of penalizing automatically, and appeals and human review further lower the effective false-flag rate. Section~\ref{sec:econ} shows that the single ratio $f/d$ decides whether the whole network can work. The judge is therefore the public research problem most worth investing in, and the job and receipt interfaces need not change as it improves.

\section{Earnings as collateral: the economics of security}
\label{sec:econ}

\paragraph{Model.} A worker earns $r$ per completed job; the payment first enters a confirmation window of $W$ days. The worker serves $\lambda$ jobs per day. Cheating on a job (running a smaller model, truncating computation, fabricating output) yields a private gain $g=\gamma r$, such as saved electricity; we conservatively assume this gain is realized immediately and can never be recovered. Job $j$ is audited with probability $p_j$ by rule~\eqref{eq:select}, and an audited cheat is judged a cheat with probability $d$. Once cheating is confirmed, the worker loses its admission bond $B$ and all pending earnings $E_j$, and its identity is quarantined. Workers are risk-neutral and we ignore discounting. A quarantined worker can rejoin under a new identity at no cost, but its collateral starts from zero. Verdicts arrive before the corresponding earnings are released.

A simple observation bounds $\gamma$: as long as honest contribution is itself profitable, the cost saved by cheating is below the payment, so $\gamma<1$. We use $\gamma=1$ as a conservative value.

\begin{proposition}[Per-job deterrence]\label{prop:rule}
If every job satisfies
\begin{equation}\label{eq:rule}
 p_j\,d\,(B+E_j)\;\geq\; g ,
\end{equation}
then no cheating strategy has positive expected excess profit over honest behaviour.
\end{proposition}
\begin{proof}
Both strategies earn the same payments; they differ only in cheating gains and forfeitures. A worker can cheat on job $j$ only if it has not yet been caught; that job brings $g$ and, with probability $p_j d$, the loss of $B+E_j$. Let $s_j$ be the probability that the worker is still uncaught at job $j$ and chooses to cheat. Expected excess profit is $\sum_j s_j\,\bigl(g-p_j d(B+E_j)\bigr)\le 0$. The future honest income lost after quarantine only makes cheating worse, and rejoining under a new identity restarts the same process, so the conclusion is unchanged.
\end{proof}

The meaning of~\eqref{eq:rule} is direct: \emph{the audit rate should be inversely proportional to the worker's current collateral}. Two corollaries follow.

\begin{corollary}[Audit load independent of throughput]\label{cor:steady}
For a worker online longer than $W$, collateral settles at $E=r\lambda W$. With $B=0$,~\eqref{eq:rule} becomes $p\geq\gamma/(d\lambda W)$, and the expected number of audits per worker per day is $p\lambda=\gamma/(dW)$, independent of $\lambda$.
\end{corollary}

With $\gamma=1$, $d=0.9$, $W=7$ days, and a safety factor $\kappa=2$, about $0.32$ replays per worker per day suffice. For workers serving $200$, $2{,}000$, and $20{,}000$ jobs per day, the lower bounds on the audit rate are $7.9\times10^{-4}$, $7.9\times10^{-5}$, and $7.9\times10^{-6}$. In other words, the cost of security is set by the number of workers, not by the volume of business: as volume grows, the audit share of total cost falls as $1/\lambda$. The busier the network, the cheaper trust becomes.

\begin{corollary}[Cold start]\label{cor:warm}
At its $j$-th job a new worker holds $E_j=jr$. Auditing with $p_j=\min\{1,\kappa\gamma/(d(j+B/r))\}$ satisfies~\eqref{eq:rule}. If the first job is always audited, the required bond is $B\geq(\gamma/d-1)^{+}\,r$, and the expected number of replays before the collateral fills is approximately $(\kappa\gamma/d)\bigl(1+\ln(d\lambda W/(\kappa\gamma))\bigr)$, growing only logarithmically with throughput.
\end{corollary}

With the parameters above, the first-job bond is only about $0.11r$, one ninth of a single payment, and a new identity consumes about $16$, $21$, and $26$ replays before its collateral fills, for $200$, $2{,}000$, and $20{,}000$ jobs per day. Because the audit rate decreases with job count rather than with time, a worker that takes a sudden burst of jobs right after joining is covered in the same way. Sybil attacks gain nothing here: every new identity starts with zero collateral and intensive auditing~\citep{douceur2002}. This lets OIM do something many staking protocols cannot: an owner with one GPU and no capital can join on equal terms.

\begin{proposition}[Feasibility depends only on the judge]\label{prop:feas}
Suppose an honest worker can tolerate losing at most a share $\varepsilon$ of its income to false flags. An audit rate satisfying both~\eqref{eq:rule} (with $B=0$) and this participation constraint exists if and only if
\[ \frac{f}{d}\;\leq\;\frac{\varepsilon}{\gamma}, \]
a condition independent of $\lambda$, $W$, and the worker's age.
\end{proposition}
\begin{proof}
An honest worker holding collateral $E$ is falsely flagged $\lambda p f$ times per day, losing $E$ each time, while earning $\lambda r$ per day. The loss share is $pfE/r\le\varepsilon$, i.e.\ $p\le\varepsilon r/(fE)$. Rule~\eqref{eq:rule} requires $p\ge\gamma r/(dE)$. Both bounds are inversely proportional to $E$, so the interval is non-empty if and only if $f/d\le\varepsilon/\gamma$.
\end{proof}

\begin{figure}[t]
\centering
\begin{tikzpicture}
\begin{axis}[
  width=0.62\textwidth, height=6.2cm, xmode=log, ymode=log,
  xmin=1, xmax=168, ymin=1e-5, ymax=1,
  xlabel={Worker age (hours; collateral fills at $W=168$\,h)}, ylabel={Audit rate $p$},
  xtick={1,6,24,72,168}, xticklabels={1,6,24,72,168},
  legend style={font=\scriptsize, at={(1.03,1)}, anchor=north west, cells={anchor=west}},
  tick label style={font=\footnotesize}, label style={font=\footnotesize}, samples=60, domain=1:168]
  \addplot[fill=green!25, draw=none, forget plot] coordinates {(1,1.3333e-02) (168,7.9365e-05) (168,7.1429e-04) (1,1.2000e-01)} --cycle;
  \addplot[blue!70!black, very thick] {min(1, 1/(0.9*2000*x/24))};
  \addlegendentry{deterrence bound, $\lambda=2{,}000$/day}
  \addplot[blue!50, thick, dashed] {min(1, 1/(0.9*200*x/24))};
  \addlegendentry{deterrence bound, $\lambda=200$/day}
  \addplot[red!70!black, very thick] {min(1, 0.01/(0.001*2000*x/24))};
  \addlegendentry{false-flag bound, $f=10^{-3}$}
  \addplot[red!50, thick, dashed] {min(1, 0.01/(0.01*2000*x/24))};
  \addlegendentry{false-flag bound, $f=10^{-2}$ (infeasible)}
\end{axis}
\end{tikzpicture}
\caption{\textbf{Feasible audit rates under earnings as collateral} ($\gamma=1$, $d=0.9$, $\varepsilon=1\%$; values generated by \texttt{evidence/model.py}). Blue lines are the lower bound required by Proposition~\ref{prop:rule}; red lines are the participation upper bound of Proposition~\ref{prop:feas} ($\lambda=2{,}000$). Both fall together as collateral grows, so the green feasible band has the same width at every worker age and is set only by the judge's $f/d$. At $f=10^{-2}$ the upper bound lies below the lower bound and no feasible audit rate exists.}
\label{fig:feas}
\end{figure}
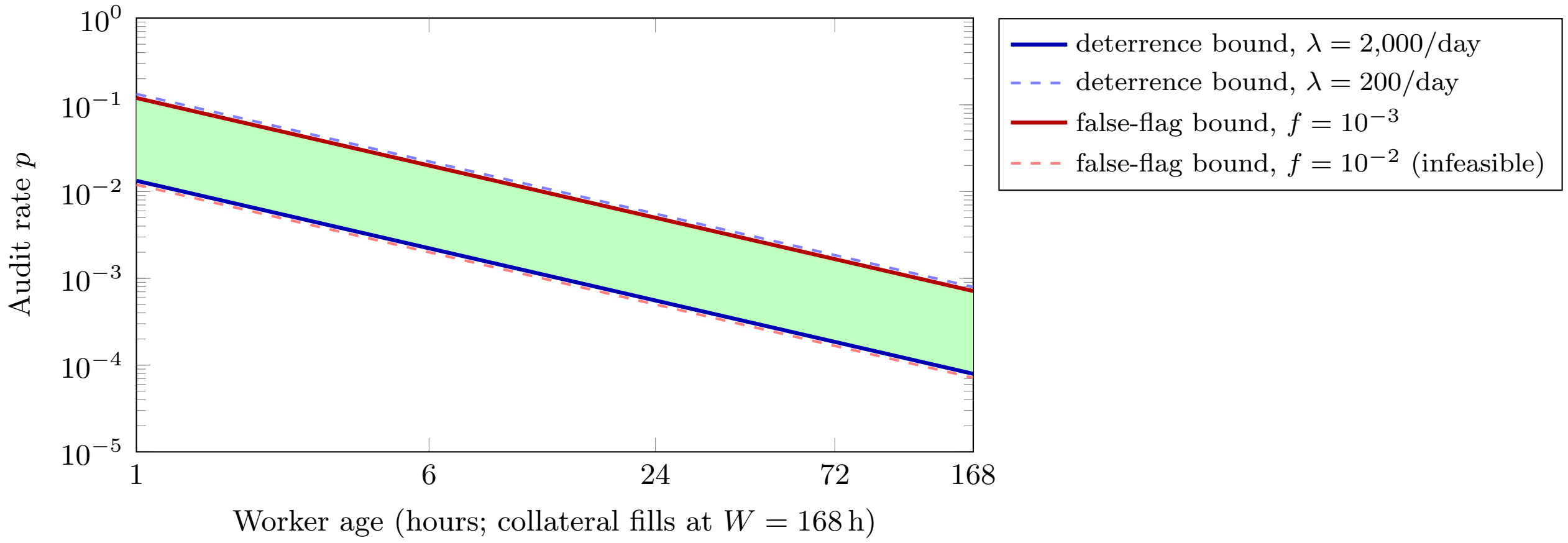

This is the paper's central result. A longer window, a larger network, or a more mature worker moves both bounds together and cannot rescue a judge whose false-flag-to-detection ratio is too high. Conversely, once the judge is good enough, the same ratio works for networks of every size and workers at every stage (Figure~\ref{fig:feas}). With $\varepsilon=1\%$, $\gamma=1$, and $d=0.9$, the requirement is $f\le0.9\%$. At $f=10^{-3}$ the two bounds differ by a factor of $9$, and the steady-state feasible interval for a worker serving $2{,}000$ jobs per day is $p\in[7.9\times10^{-5},\,7.1\times10^{-4}]$; at $f=10^{-2}$ no feasible audit rate exists. This turns a vague question of trust into a clear, measurable engineering target that many people can work on: lower the false-flag rate of inference audits.

\paragraph{The window is a dial.} Owners want to be paid early; the network wants thicker collateral. Corollary~\ref{cor:steady} gives the trade-off: at a fixed audit rate, deterrence requires $W\geq\gamma/(pd\lambda)$. With $p=10^{-3}$, $d=0.9$, and $\lambda=2{,}000$, about half a day suffices; a 7-day window leaves more than a tenfold margin, which can lower the audit rate further, accommodate low-throughput workers, or leave time for human review.

\paragraph{Preconditions.} Collateral must have real value to the worker, i.e.\ earnings can be redeemed for service or money; the settler must release payments and enforce forfeitures as agreed; verdicts must arrive within the window; and auditors must not collude with workers. The first is a matter of economic design (Section~\ref{sec:open}, Q4); the other three are exactly what the path in Section~\ref{sec:path} moves, step by step, from trusting the operator to public verification. More general cases are discussed in Section~\ref{sec:open}.

\section{How the network starts and grows}
\label{sec:growth}

\paragraph{The first market.} An open network need not compete head-on with large data centers from day one. Its first market is jobs whose models fit on a single consumer GPU, that tolerate latency of seconds, and whose data may be processed externally: batch text processing and data labelling, document summarization and translation, background subtasks run by agents, and the many repeated calls of development and evaluation. Such jobs value price and total throughput over millisecond response, and they are exactly what idle devices do best. As open-weight models keep improving and quantization and offloading keep advancing~\citep{powerinfer2023,flexgen2023}, the boundary of this market will widen naturally.

\paragraph{Why owners join.} The hardware cost of an idle GPU is already paid; the marginal cost of contributing is mostly electricity. Taking \Bee{}'s published prices as a reference (80 credits per million output tokens, USD 0.001 per credit, a 70\% contributor share), an owner earns USD 0.056 per million tokens. \emph{Assuming} a 350\,W card and electricity at USD 0.08 per kWh, electricity costs USD 0.078 per million tokens at a single-stream 100 tokens/s, falling to USD 0.019 and 0.005 when batching reaches 400 and 1{,}600 tokens/s. The conclusion is clear: an owner's income is decided not by the card itself but by whether it can be kept busy in batches~\citep{vllm2023}. This defines the gateway's core task as \emph{aggregating demand}: concentrating jobs on online workers that already have the right model loaded, so that every card runs full batches.

\paragraph{Two network effects that grow with scale.} The first is a supply-demand flywheel: more owners bring more models, regions, and hours, so queues shorten and prices fall, which brings more demand; more demand raises each card's utilization and batching efficiency, which raises owners' income, which brings more owners. The second is a security flywheel unique to OIM: by Corollary~\ref{cor:steady}, each worker's daily audit load is constant, so the further the first flywheel pushes each card's volume, the smaller the audit share of cost. Conventional trust institutions need more reviewers as they grow; in OIM the unit cost of trust falls with scale.

\paragraph{Availability.} Home workers sleep and disconnect, and loading a model takes time. The gateway spreads replicas of each model across distinct power and network failure domains and keeps a small amount of stable seed capacity as a fallback. If replicas are available independently with probabilities $a_i$, the probability that all are down is $\prod_i(1-a_i)$; three home replicas that are each only 90\% available would in principle give 99.9\% availability. Real failures are correlated, so the network should publish completion rates, retries, and tail latency by failure domain, and honour its service promises by measurement rather than by extrapolation.

\paragraph{Privacy.} The owner of an ordinary external worker can read the prompts it receives, so OIM divides data into tiers and lets each tier promise only what it can deliver. Public-tier jobs go to external workers only with the requester's explicit consent; sensitive jobs stay on devices the requester trusts or on the operator's seed capacity; and a confidential tier can open once trusted execution environments and remote attestation mature under a clear threat model~\citep{costan2016sgx}. Drawing boundaries honestly is how an open network earns trust.

\section{From distributed machines to distributed trust}
\label{sec:path}

Bitcoin handed every role to the network at once. An inference network need not, and should not: a useful service needs someone answerable for delivery first. OIM takes another route: open the supply of compute first, then hand over accounting, adjudication, and settlement one at a time. Decentralization can be split into four powers: \emph{who owns the compute, who writes the history, who judges the work, and who releases the value}. Each stage transfers one of them and publishes evidence anyone can check (Table~\ref{tab:path}).

\begin{table}[h]
\centering\small
\begin{tabularx}{\textwidth}{@{}p{24mm}p{20mm}YY@{}}
\toprule
Stage & Power handed over & How & Evidence to publish\\
\midrule
S1 External supply & Owning compute & The operator offers one interface, dispatches consented public-tier jobs to admitted independent GPUs, and is answerable for delivery and payment & Share of jobs served by independent owners; full cost and failure rate per successful job\\
S2 Checkable accounts & Writing history & Every job has a pre-dispatch commitment and a co-signed receipt, written to an append-only log observed by independent monitors & Re-checkable receipt samples; detection of omitted or conflicting records\\
S3 Independent audit & Judging work & Public randomness selects audits; auditors outside the gateway sign appealable verdicts under audit contracts & $d$ and $f$ by model and hardware; audit cost and dispute resolution time\\
S4 Portable settlement & Releasing value & Multiple operators accept one set of quotes, receipts, verdicts, and forfeiture rules, strengthened where useful by a shared ledger or on-chain anchoring & Cross-operator payouts and final outcomes of disputes\\
\bottomrule
\end{tabularx}
\caption{The trust handover path. The four dimensions are independent; progress on one does not complete the others, and each stage can be started and checked on its own.}
\label{tab:path}
\end{table}

What matters about this path is that \emph{it can start now}. S1 needs no new cryptography, only an operator willing to answer for delivery and owners willing to contribute GPUs; \Bee{} is one such early instance (Appendix~\ref{app:beenet}). Each stage reduces participants' dependence on the operator, and by S4 the exit of any single operator no longer means the end of the network. Blockchains and tokens are optional tools on this path, not prerequisites: a shared ledger is adopted only where it actually makes settlement more enforceable.

\section{Security model and building agenda}
\label{sec:open}

The results of Section~\ref{sec:econ} rest on a simplified model. We collect its assumptions and open problems here; they are also where the network most needs contributors.
\begin{enumerate}[label=\textbf{Q\arabic*.}]
\item \textbf{Judges on heterogeneous hardware.} Proposition~\ref{prop:feas} reduces everything to $f/d$. On uniform hardware with greedy decoding, replay comparison can have a very low false-flag rate; across GPU models, batches, and sampled decoding the rate must still be measured model by model~\citep{atil2024nondet,he2025nondet}. Activation fingerprints and distributional comparison are the most promising directions~\citep{toploc2025,difr2025}.
\item \textbf{Collusion and repeated games.} The model assumes auditors do not collude with workers and that workers are risk-neutral and do not discount. Repeated-game analysis shows that single-round conditions can overstate security when penalties shrink with history~\citep{shang2026repeated}; forfeiture under earnings as collateral depends on history, and its full repeated-game analysis remains open. Cross-checking by several independent auditors is a natural mitigation.
\item \textbf{Bounding the cheating gain.} $g$ counts only saved compute. If a worker can profit elsewhere, for example by reselling prompts, $\gamma$ may exceed $1$; this is why the data tiers of Section~\ref{sec:growth} send only public-tier jobs to external workers.
\item \textbf{Value of collateral.} Non-transferable service credits avoid speculation, but their value depends on whether a worker can use them. Redemption paths and owners' real valuation of credits need to be measured.
\item \textbf{Cost of false flags.} Proposition~\ref{prop:feas} counts only forfeited collateral; a falsely flagged honest worker also loses time and reputation. Appeals, human review, and abstention keep the effective $f$ well below a judge's raw false-flag rate, and their effect should also be reported publicly.
\item \textbf{Operator honesty.} Until S2 and S3 are complete, accounts and verdicts still depend on the operator. Every stage of the path uses public evidence as its milestone precisely so that this dependence can be checked and replaced step by step.
\end{enumerate}

\section{Related work}
\label{sec:related}

\paragraph{Blockchains and useful work.} Bitcoin maintains a public ledger with proof-of-work~\citep{nakamoto2008}; proofs of useful work try to make that work useful as well~\citep{ball2017pouw,ofelimos2022}. OIM does not use inference to maintain global consensus; it settles individually purchased jobs and therefore needs only honesty in expectation. Decentralized physical infrastructure networks (DePIN) and Bittensor organize hardware and models with token incentives~\citep{lin2024depin,ballandies2023depin,bittensor2020}; OIM does not presuppose issuing a token.

\paragraph{Verifiable inference.} Existing protocols secure inference with replay, sampling, staking, or proofs~\citep{verillm2025,posp2024,shang2026repeated}, and fingerprinting and zero-knowledge methods lower the cost of verification~\citep{toploc2025,difr2025,zkllm2024}. Staking schemes require workers to lock capital in advance; earnings as collateral lets collateral arise endogenously from a worker's own work, and gives closed-form conditions for audit load, cold-start cost, and feasibility.

\paragraph{Spot checks, optimistic execution, and transparency logs.} Sabotage tolerance in volunteer computing and uncheatable distributed computation pioneered random spot checks~\citep{sarmenta2002,golle2001,kondo2007}; TrueBit, Arbitrum, and opML achieve optimistic verification with challenge periods~\citep{truebit2019,arbitrum2018,opml2024}; Certificate Transparency provides the engineering pattern for auditable logs~\citep{laurie2013ct}. OIM combines these into one complete path from dispatch to settlement for consumer GPUs.

\paragraph{Distributed inference and training.} Petals, collaborative inference, and heterogeneous serving systems study model serving on dispersed GPUs~\citep{petals2022,borzunov2023,hexgen2024,parallax2025}; progress in distributed training shows that dispersed devices can also take part in heavier work~\citep{diloco2023,intellect1}. These systems address how to compute; OIM addresses how strangers can trust that the computation was done.

\section{Conclusion}
\label{sec:conclusion}

Bitcoin answered a question: can strangers jointly maintain a ledger without a trusted intermediary? This paper addresses its counterpart in the age of AI: can strangers organize the idle compute in their hands into inference services they can trust one another to deliver? Our answer is yes, and without waiting for perfect cryptography. As long as audits are unpredictable, earnings serve as collateral during the confirmation window, and the judge's false-flag rate is sufficiently below its detection rate, cheating does not pay; and the cost of maintaining this is a fraction of a replay per worker per day, spread ever thinner as the network grows.

This means the next AI compute center need not be a building but a protocol, whose capacity is set not by one organization's capital but by how many people choose to join. A student's graphics card, a design studio's workstation at night, the idle hours of a school computer lab can all become part of it and be rewarded under public, checkable rules.

None of this requires one institution to solve every problem first. Owners can contribute compute, requesters can bring jobs, model developers can write audit contracts, researchers can lower the false-flag rate of judges, and independent monitors can check the accounts. Every job that is honestly settled moves the open compute network one step forward. We invite everyone who owns compute, needs compute, or cares who controls compute to help build it.


\appendix
\section{An early instance: BeeNet}
\label{app:beenet}

\Bee{} is a live inference network that issues no token and can be viewed as an instance of stage S1. Public pages archived by the authors on 18 October 2026 describe an OpenAI-compatible inference interface, contributor admission, sensitivity-tiered dispatch, receipts signed by both sides that contain only hashes, replay audits run by the operator, a 7-day earnings confirmation period, and a 70\% contributor share computed in internal credits~\citep{beenetweb}. The combination of a confirmation period and audits is one engineering form of earnings as collateral. All of this comes from operator documentation; we could not independently verify the internal implementation.

On the same day we sent one smoke-test request and 31 experimental requests through the public trial flow, and all returned successfully. Sixteen greedy-decoding requests across two models produced output identical word for word within each model, and a pair of fixed-seed sampled requests also matched. This is consistent with the expectation that replay comparison can serve as a judge on uniform hardware with deterministic settings, but the sample is small and cannot be used to estimate a false-flag rate. All requests showed the same \texttt{rtx-4090/cuda} cluster label, so these observations do not speak to the behaviour of independent owners' devices, cheating detection rates, or network economics. Scripts and raw responses are provided as ancillary files.
\end{document}